\documentclass{Interspeech2024}

\interspeechcameraready

\usepackage{subcaption}
\usepackage{tikz}
\usetikzlibrary{positioning}
\usetikzlibrary{calc}
\usepackage{fontawesome5}
\usepackage{multirow}

\usepackage{tikz}
\usepackage{everypage}

\newcommand\BackgroundText{%
  \begin{tikzpicture}[remember picture,overlay]
    \node [rotate=0, scale=1.3, text opacity=1.0, color=gray] at (current page.south west) [anchor=south west, xshift=0.7cm, yshift=1.9cm] {accepted at Interspeech 2024};
  \end{tikzpicture}
}

\AddEverypageHook{\BackgroundText}

\title{Controlling Emotion in Text-to-Speech with Natural Language Prompts}
\name{Thomas Bott, Florian Lux, Ngoc Thang Vu}

\address{University of Stuttgart, Germany}
\email{[thomas.bott, florian.lux, thang.vu]@ims.uni-stuttgart.de}

\begin{document}

\maketitle
 
\begin{abstract}
In recent years, prompting has quickly become one of the standard ways of steering the outputs of generative machine learning models, due to its intuitive use of natural language. In this work, we propose a system conditioned on embeddings derived from an emotionally rich text that serves as prompt. Thereby, a joint representation of speaker and prompt embeddings is integrated at several points within a transformer-based architecture. Our approach is trained on merged emotional speech and text datasets and varies prompts in each training iteration to increase the generalization capabilities of the model. Objective and subjective evaluation results demonstrate the ability of the conditioned synthesis system to accurately transfer the emotions present in a prompt to speech. At the same time, precise tractability of speaker identities as well as overall high speech quality and intelligibility are maintained.
\end{abstract}

\section{Introduction}

With rapid advancements in text-to-speech (TTS) systems in recent years \cite{Ren2020, kharitonov2023speak, wang2023neural}, speech can be synthesized with naturalness and intelligibility comparable to human speakers \cite{Liu2021a, tan2024naturalspeech}.
However, the one-to-many-mapping problem remains as one of the fundamental challenges. This refers to the fact that for a given input text there are infinitely many valid realizations which can differ in their prosody, including speaking style, intonation, stress or rhythm.
A frequently used approach for mitigating this problem is to enrich the input side, i.e. the text to be encoded, with auxiliary prosodic information to alleviate the mismatch in the mapping \cite{Ren2020, Lancucki2021}. These additional prosodic inputs can often be controlled during inference time. Many previous approaches rely on reference audios for transferring the desired speaking style \cite{SkerryRyan2018, Wang2018, Yan2021, Casanova2022, lux2023exact}. However, such methods oblige users to provide reference speech with the desired criteria during inference, which may not always be available. Addressing this issue, recent work has focused on using natural language descriptions to guide prosodic aspects in TTS systems, trained on speech datasets augmented with style descriptions \cite{Kim2021, Shin2022, Guo2023, Yang2023, Liu2023}. Style tagging TTS \cite{Kim2021} introduces a specialized loss, allowing to either provide reference speech or style tag during inference. PromptTTS \cite{Guo2023} fine-tunes style embeddings on pre-defined labels such as gender, pitch, speaking speed, volume and emotion. PromptStyle \cite{Liu2023} and InstructTTS \cite{Yang2023} introduce a cross-modal style encoder which learns a shared embedding space of prompt and style embedding from speech. These approaches however require datasets with style descriptions which are expensive to create. Moreover, manually provided style descriptions are limited, as they typically follow similar patterns. PromptTTS 2 \cite{Leng2023} tries to overcome this issue by labeling voice characteristics such as gender and speed from audio and generating descriptive prompts based on these properties automatically, which however limits the granularity of their control. Since emotional states are one of the most obvious aspects that can be expressed by varying prosodic features \cite{Leentjens1998, Sauter2010, Pell2011}, emotional TTS is an important subfield within controllable TTS. In this vein, \cite{Tu2022} automatically extract prompts from an emotional text dataset and match them to speech samples annotated with emotion labels.

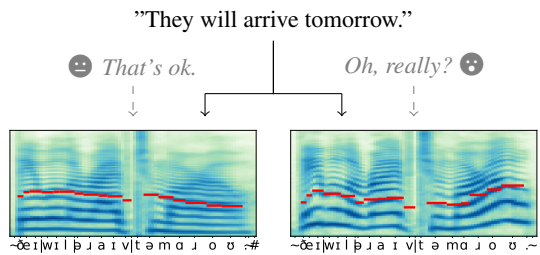
\begin{figure}
    \centering
    \begin{tikzpicture}
        % Include first image
        \node[anchor=south west,inner sep=0] (image1) at (0,0) {\includegraphics[width=0.45\columnwidth]{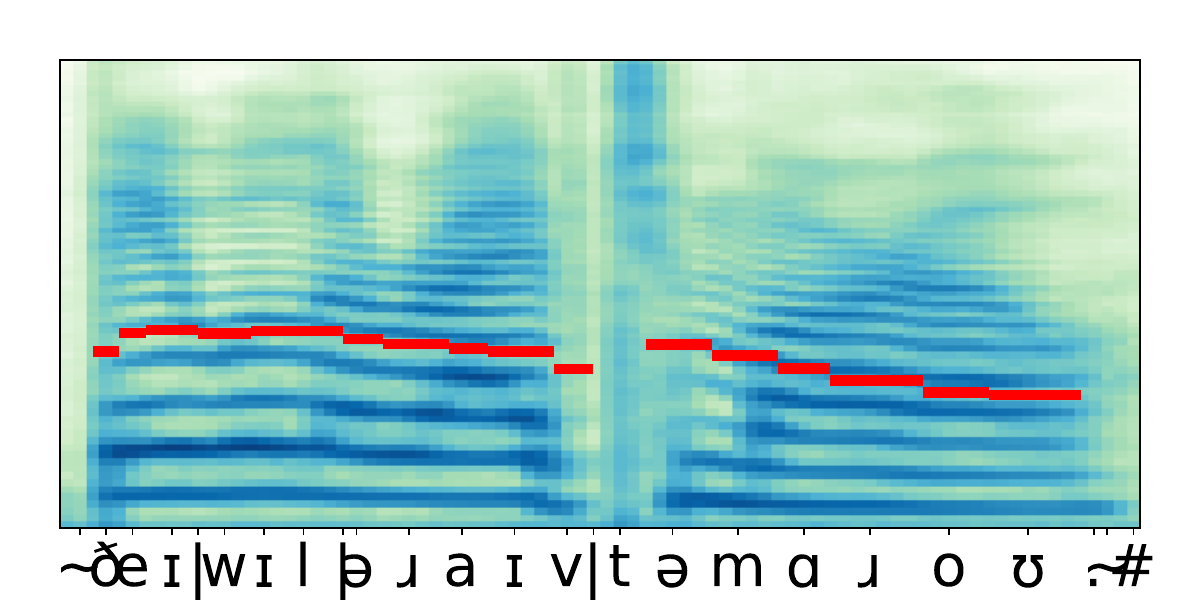}};
        % Add text and arrow
        \node[above=0.4cm of image1,align=center, text=black!50] (text1) {\faMeh \hspace{0.05cm} \textit{That's ok.}};
        \draw[->, dashed, black!50] (text1) -- (image1);
        
        % Include second image
        \node[anchor=south west,inner sep=0] (image2) at ([xshift=0.01\columnwidth]image1.south east) {\includegraphics[width=0.45\columnwidth]{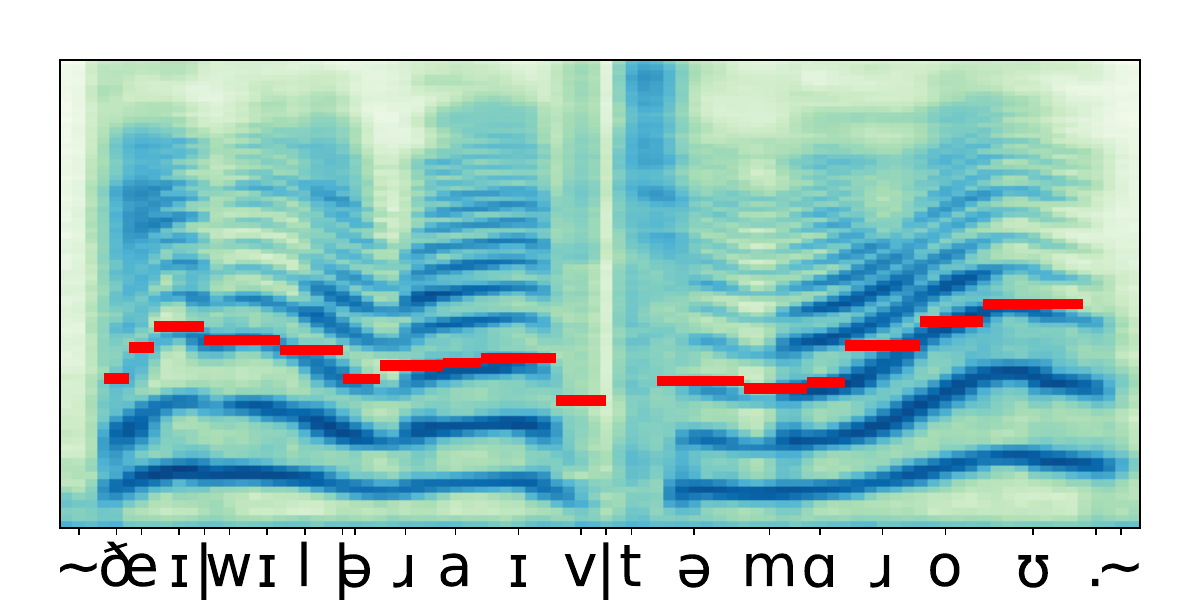}};
        % Add text and arrow
        \node[above=0.4cm of image2,align=center, text=black!50] (text2) {\textit{Oh, really?} \faSurprise};
        \draw[->, dashed, black!50] (text2) -- (image2);
        
        % Add common text and arrows
        \node[above=1cm of $(image1.north)!0.5!(image2.north)$,align=center] (commonText) {"They will arrive tomorrow."};
        % Draw lines and arrows
        \draw (commonText) -- ([yshift=0.3cm]commonText |- image1.north);
        \draw (commonText) -- ([yshift=0.3cm]commonText |- image1.north) -- ++(-0.9cm,0) coordinate (leftLineEndpoint);
        \draw (commonText) -- ([yshift=0.3cm]commonText |- image1.north) -- ++(0.9cm,0) coordinate (rightLineEndpoint);

        \draw[->] (leftLineEndpoint) -- ++(0,-0.3cm);
        \draw[->] (rightLineEndpoint) -- ++(0,-0.3cm);
    \end{tikzpicture}
    \caption{Spectrograms with pitch contour for the same text, synthesized by our proposed system given two different emotional prompts. On the left the underlying emotion is neutral ("That's ok.") and on the right it is surprise ("Oh, really?").}
    \label{fig:teaser}
\end{figure}

Our approach follows a similar strategy, combining publicly available emotional speech and text datasets and obtaining strong dependency between prosodic properties of audio and prompts. Furthermore, during each training iteration, prompts are selected randomly from a large pool, which increases the generalization capabilities of the TTS system and reduces the risk of learning connections that are too specific. In contrast to \cite{Tu2022}, who model speaker identities within the prompt, our method effectively combines prompt and speaker embeddings, allowing for precise prosodic and timbral controllability.

\begin{figure*}[ht]
    \centering
    \includegraphics[width=0.7\textwidth, clip, trim=1.1cm 3.7cm 1.7cm 2.9cm]{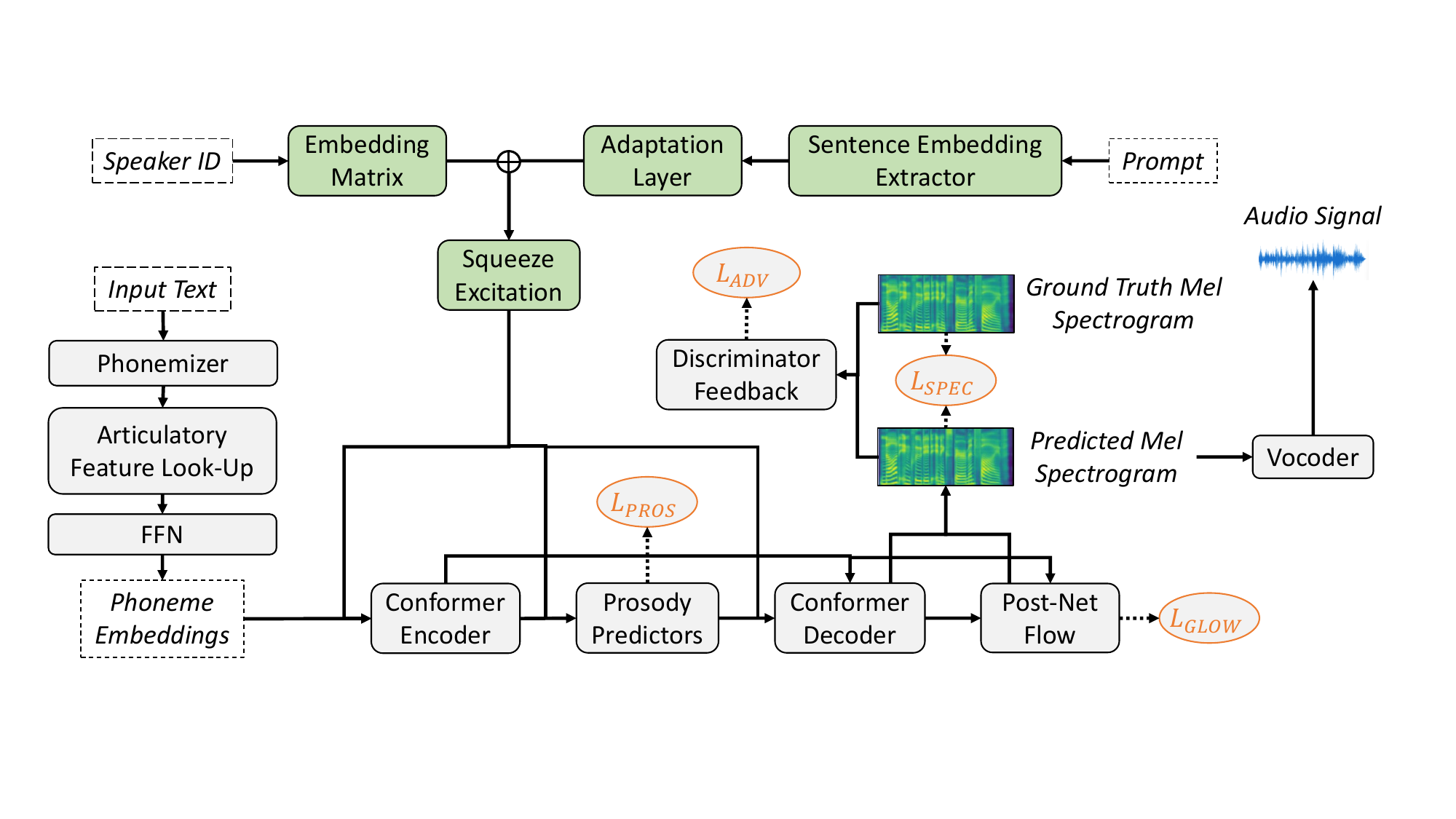} 
    \caption{Architecture of the prompt conditioned TTS system. Green components handle the integration of speaker and prompt embedding. $+$ indicates concatenation. The loss functions with which the components in this system are optimized are marked in orange.}
    \label{fig:proposed}
\end{figure*}

Summarizing our contributions, we propose 1) an architecture that allows for separate modeling of a speaker's voice and the prosody of an utterance, using a natural language prompt for the latter, 2) a training strategy to learn a strongly generalized prompt conditioning, and 3) a pipeline that allows users to generate speech with fitting prosody without manually selecting the emotion by simply using the text to be read as the prompt. We evaluate our contributions objectively and subjectively, finding that the emotion present in the prompt can be accurately transferred to speech while maintaining precise tractability of speaker identities and high speech quality.
All of our code and models are available under an open source license\footnote{\url{https://github.com/DigitalPhonetics/IMS-Toucan/tree/ToucanTTS_Prompting}}.

\section{Resources}

\subsection{Speech Datasets}
We select two datasets without emotion annotation, LJSpeech\footnote{\url{https://keithito.com/LJ-Speech-Dataset}} (single speaker, 24 hours) and LibriTTS-R \cite{Koizumi2023a, Zen2019} (2,456 speakers, 585 hours), as well as three datasets which contain an emotion label along with each utterance: Emotional Speech Database (ESD) \cite{Zhou2021a}, Ryerson Audio-Visual Database of Emotional Speech and Song (RAVDESS) \cite{Livingstone2018} and Toronto Emotional Speech Set (TESS) \cite{PichoraFuller2020}.
All used datasets are publicly available, an overview of the properties of the emotional speech datasets can be seen in Table \ref{tab:speech_datasets_emo}.

\begin{table}
\caption{Properties of the emotional speech datasets used for training. Each dataset is annotated with emotion categories.}
	\begin{center}
		\begin{tabular}{l ccc}
			\toprule
			& ESD & RAVDESS & TESS  \\
			\midrule
			Speakers & 10 & 24 & 2  \\
			Emotion Categories & 5 & 8 & 7 \\
			Utterances per Emotion & 350 & 8 & 200 \\
			Utterances in Total & 17,500 & 1,440 & 2,800 \\
			\bottomrule
		\end{tabular}
	\end{center}
	\label{tab:speech_datasets_emo}
\end{table}

\subsection{Text Datasets}
In order to obtain non-descriptive emotional prompts, we extract sentences from text datasets labels with emotions. For training, a subset of the Yelp open dataset \cite{Asghar2016} is used, which contains 650k reviews for businesses and restaurants that contain highly emotional text from various different authors. We extract 10k prompts for each emotion category and confirm the emotion label with an auxiliary emotion classification model \cite{Hartmann2022}. For the evaluation, we extract prompts that will remain unseen during training from DailyDialog \cite{Li2017} (containing 13,118 annotated sentences) and Tales-Emotion \cite{Alm2005, Alm2005a} (containing 15,292 annotated sentences). From both datasets, 50 prompts per emotion category are extracted while ensuring quality with help of the emotion classification model. The conversational nature of the sentences in DailyDialog and the story telling style in Tales-Emotion demonstrate the usefulness of the TTS model for both dialog systems and reading stories. All used datasets are publicly available.

\section{Methods}

\subsection{Prompt Conditioned TTS Architecture}

An overview of the system is shown in Figure \ref{fig:proposed}. We base our implementation on the IMS Toucan toolkit \cite{Lux2021, Lux2023a} and extend it in order to condition the model on the emotional content of the textual prompt. The input text is converted into a sequence of phonemes using a phonemizer\footnote{\url{https://github.com/bootphon/phonemizer}} with eSpeak-NG\footnote{\url{https://github.com/espeak-ng/espeak-ng}} as back-end. Each phoneme is further transformed into an articulatory feature vector, following \cite{Lux2022}. Spectrogram frames are generated by a FastSpeech-2-like system \cite{Ren2020} comprising a Conformer encoder and decoder \cite{Gulati2020} as well as prosody predictors for duration and pitch and energy per phoneme as in FastPitch \cite{Lancucki2021} using a small self-contained aligner following \cite{aligner}. As proposed in PortaSpeech \cite{Ren2021} a normalizing-flow-based post-net is used to improve details in high frequencies. Finally, the system is trained with discriminator feedback coming from an adversarial network optimized to distinguish real and generated spectrograms as proposed in \cite{Sani2023}.

Natural language prompts are fed into a sentence embedding extractor based on a DistilRoBERTa \cite{Sanh2019} model fine-tuned on the task of emotion classification \cite{Hartmann2022}. Embeddings are obtained from the 756 dimensional hidden representation of the \textit{[CLS]} token. Since the emotion classification is based on the embedding of this token, it is expected to effectively capture relevant information about the emotional content of the input. These prompt embeddings are further passed through a linear layer to enable them to adapt for TTS purposes, as the prompt encoder is not updated during TTS training. On the contrary, speaker embeddings are obtained from an embedding matrix, which is updated jointly during TTS training to capture the different speaker identities. To allow for zero-shot voice adaptation, a pretrained speaker embedding function could be used instead \cite{Jia2018}, which we choose to leave out for simplicity in this study. Prompt and speaker embeddings are concatenated and passed through a squeeze and excitation block \cite{Hu2018}. This component models inter-dependencies between features of both sources and projects into the hidden dimensionality of the system. 
The use of the squeeze and excitation block is motivated by an internal pilot study, in which we compare the effectiveness of using various forms of conditioning mechanisms for this, such as concatenation followed by projection \cite{Jia2018}, addition \cite{Lancucki2021}, conditional layernorm \cite{Wu2022AdaSpeech4A}, and the squeeze and excitation block \cite{Hu2018}. Despite the differences being minor, we decided to go forwards with the squeeze and excitation block due to its slightly better performance in picking up fine nuances in a conditioning signal perceptually. The output of this block is a representation that contains information about the speaker identity and semantics of the prompt. This representation is integrated in the TTS system's pipeline by providing it as auxiliary input to the encoder and decoder as well as to the prosody predictors. In these places, it is integrated using conditional layernorm \cite{Wu2022AdaSpeech4A}, which is proven to work well in a TTS pipeline \cite{Wu2022AdaSpeech4A, Lux2023a}. Adding the conditioning signals in multiple places is motivated by StyleTTS \cite{Li2022} who argue that a model quickly forgets about conditioning signals and needs to be reminded of them for more accurate conditioning.
Finally, the spectrograms are converted to waveforms using the HiFi-GAN \cite{Kong2020} generator with the Avocodo \cite{Bak2022} discriminator set. During inference, this pipeline achieves a real-time-factor of 0.07 on a Nvidia GeForce RTX 2080 Ti GPU and 0.16 on a AMD EPYC 7542 CPU without the use of batching.

\begin{table}
\caption{Results for speaker similarity in terms of cosine similarity between speaker embeddings of ground truth and synthesized audio across all speaker ids of ESD.}
	\begin{center}
		\begin{tabular}{l c c}
			\toprule
            System & $\mu$ & $\sigma$ \\
            \midrule
			Baseline  & 0.950 & $\pm$0.0044 \\
            Prompt Conditioned  & 0.953 & $\pm$0.0025 \\
            EmoSpeech & 0.861 & $\pm$0.0115 \\
			\bottomrule
		\end{tabular}
	\end{center}
	\label{tab:speaker_similarity}
\end{table}

\subsection{Prompt Inducing Training Procedure}
The training of the TTS system is carried out by curriculum learning with two stages. Although conditioning prompts are still used in the first stage, its main purpose is to obtain a robust and high quality system. Therefore this stage includes LJSpeech and LibriTTS-R in addition to the emotional speech datasets. The large amount of training samples and huge variety of speakers is beneficial for a high speech quality and makes the system more robust against mispronunciations. Since LJSpeech and LibriTTS-R do not contain emotion labels, prompt embeddings are extracted from the corresponding utterances themselves. During the second stage, the model is trained using only the emotional speech datasets, allowing it to focus on learning the connection between prompt embeddings and speech emotion.
For each training sample, a prompt embedding is selected randomly from the 10k available ones based on the emotion label. This ensures a high correspondence between prompt and speech emotion and further has the advantage that a large number of different prompts is seen which reduces the risk of overfitting and increases the generalization capabilities of the system such that during inference arbitrary prompts can be used. The whole system is trained for 120k steps during the first stage and an additional 80k steps in the second stage on a single Nvidia GeForce RTX A6000 GPU.

\begin{table}
\caption{Cramér's V values indicating association strength between predicted emotions and underlying labels. For Prompt Conditioned Same input text is the same as prompt while Prompt Conditioned Other uses a different prompt from the text. All values are statistically significant with respect to the $\chi^2$-test between predicted and underlying emotion labels with $\alpha=0.005$.}
	\begin{center}
		\begin{tabular}{l c}
			\toprule
            System & Cramér's V\\
            \midrule
			Ground Truth & 0.85\\
            Baseline & 0.06\\
            Prompt Conditioned Same & 0.80\\
            Prompt Conditioned Other & 0.80\\
            EmoSpeech & 0.96 \\
			\bottomrule
		\end{tabular}
	\end{center}
	\label{tab:cramer_emotion_recognition}
\end{table}

\begin{table}
\caption{Mean opinion score on a 5-point scale indicating naturalness, fluency and intelligibility as perceived by human raters.}
	\begin{center}
		\begin{tabular}{l c c}
			\toprule
            System & MOS & $\sigma$ \\
            \midrule
           Ground Truth  & 3.95 & $\pm$0.42 \\
			Baseline  & 3.30 & $\pm$0.45 \\
           Prompt Conditioned  & 3.37 & $\pm$0.31 \\
			\bottomrule
		\end{tabular}
	\end{center}
	\label{tab:mos}
\end{table}

\begin{figure*}[ht]
    \centering
    \includegraphics[width=\textwidth, clip, trim=6.9cm 4.4cm 1.6cm 4.9cm]{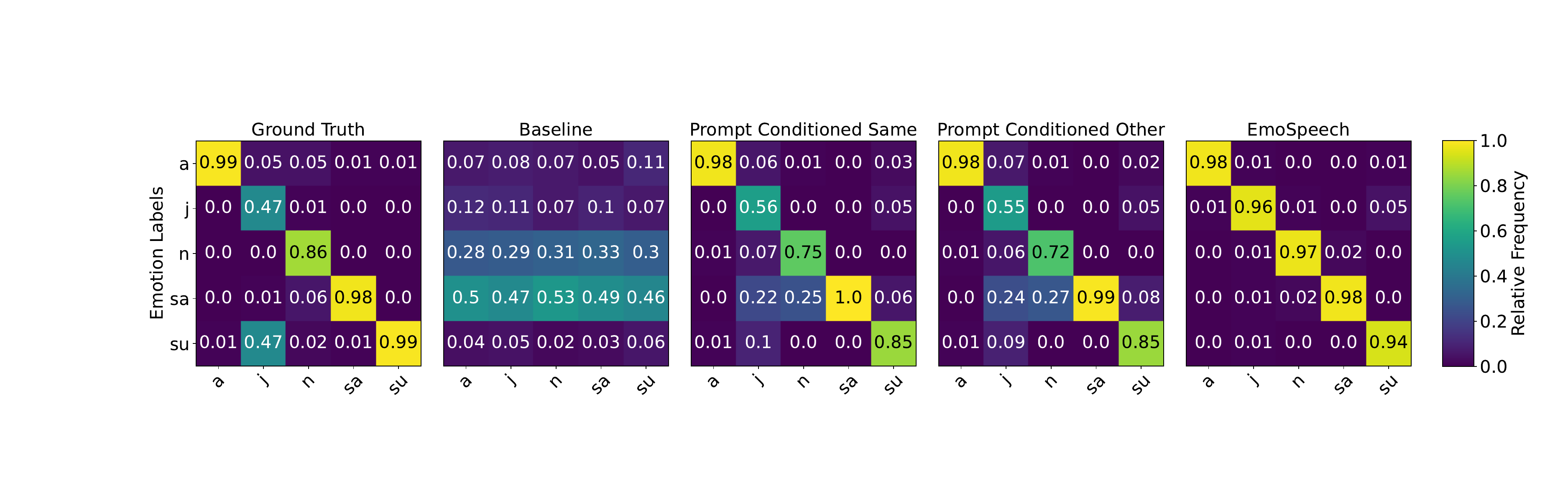}
    \caption{Results of speech emotion recognition in terms of relative frequency for predicted emotion labels opposed to underlying ones. For Prompt Conditioned Same input text is the same as prompt while Prompt Conditioned Other uses different prompts. Emotion labels are abbreviated as follows: a(nger), j(oy), n(eutral), sa(dness), su(rprise).}
    \label{fig:emo_rec}
\end{figure*}

\section{Experiments}

\subsection{Experimental Setup}

We compare the proposed TTS system to a baseline following the exact same architecture except for the missing conditioning on prompt embeddings. 
Additionally we include EmoSpeech \cite{diatlova2023emospeech} \footnote{\url{https://github.com/deepvk/emospeech}} in our objective evaluation which conditions a FastSpeech 2 architecture on the discrete emotion labels of ESD.
For our prompt conditioned system, the test sentences from the emotional text datasets are synthesized using the sentences themselves as prompts as well as using sentences annotated with a different emotion as prompts. This allows us to assess if the generated speech emotion relies on the provided prompt embedding throughout all experiments we are conducting. We also pass all ground truth speech samples through the vocoder of the TTS system, in order to allow a fair comparison to synthesized speech. Speaker identities from ESD are used for evaluation purposes, including the emotion categories \textit{anger}, \textit{joy}, \textit{neutral}, \textit{sadness} and \textit{surprise}.

% only included selected results, everything is too much
% objective: speaker similarity, emotion recognition
% subjective: mos overall, A/B preference, prosodic similarity

\subsection{Objective Evaluation}

\subsubsection{Multi-Speaker Capabilities}
We calculate the speaker similarity as the cosine similarity between speaker embeddings of ground truth speech samples and synthesized ones. Thereby speaker embeddings are extracted using a pre-trained speaker verification model \cite{Snyder2018}. The results in Table \ref{tab:speaker_similarity} show a high overall speaker similarity across all speakers of ESD, demonstrating that speaker identities are almost perfectly preserved during synthesis and not affected by the integration of prompt embeddings. Comparing our results to EmoSpeech, both our proposed system and our baseline perform significantly better. This is likely caused by the multi-speaker training phase in our curriculum learning procedure.

\subsubsection{Prosody Controllability}
We use an auxiliary speech emotion recognition model \cite{Ravanelli2021} trained on ESD to predict emotion labels of synthesized speech and compare those to the ground truth labels of the applied prompts. The confusion matrices in Figure \ref{fig:emo_rec} illustrate the relative frequencies of predicted emotion labels opposed to underlying ones. Furthermore, as a measure of association strength between emotion labels, Cramér's V \cite{Cramer1999} is calculated with results shown in Table \ref{tab:cramer_emotion_recognition}. The emotion recognition model achieves an overall high accuracy and high association strength for ground truth speech, demonstrating that the emotion can generally be recognized reliably. Considering this, the strong alignment between underlying and predicted emotion labels for the prompt conditioned system indicates that the emotional content of the prompt is accurately transferred to speech. Moreover, the speech prosody exclusively relies on the provided prompt and is not influenced by the input text of the synthesized utterance as revealed by the high accuracy when combining prompts and input texts that stem from different emotion categories (``Prompt Conditioned Other''). In contrast, for the baseline, the predicted emotion categories are mostly sadness and neutral, showing that there is hardly any prosodic variation in generated speech irrespective of the emotional content of the input text. These observations are further confirmed by the Cramér’s V values for the prompt conditioned system which are on par with ground truth following a student's t-test with $\alpha=0.005$.
EmoSpeech yields very strong results, outperforming even the ground truth. It is, however, restricted to discrete emotion labels, while our system captures a continuous space that doesn't require the manual selection of an appropriate emotion. This offers a great advantage over the state-of-the art of specialized systems, such as EmoSpeech, at the cost of a slightly degraded emotion accuracy. 

\subsection{Subjective Evaluation}
Due to EmoSpeech's large variance in quality and intelligibility, which we noticed in a small pilot study, we chose to exclude it from the subjective evaluation, to prevent ceiling effects. Hence we compare our proposed system only with the baseline and human recordings in the following. We conducted a listening study with 82 participants, utilizing test sentences generated with both a female and a male speaker identity from ESD and varying prompts.

\subsubsection{Speech Quality}
We ask participants to rate speech quality on a 5 point scale, considering naturalness, fluency and intelligibility. The results of this mean opinion score (MOS) study based on 656 ratings (Table \ref{tab:mos}) indicate that synthesized speech from both the baseline as well as the proposed system is only slightly but significantly degraded compared to ground truth speech, yet not significantly different from one another (following a student's t-test with $\alpha=0.005$). We conclude that the addition of prompt conditioning does not hinder the perceived naturalness of the TTS system.

%\subsubsection{Emotional Preference Test}
%Participants are further asked to listen to synthesized speech from the baseline and the proposed system and then select the preferred one with respect to its prosodic appropriateness to the emotional content of the utterance. Figure \ref{fig:pref_total} shows that in the 328 preference ratings we collected, the proposed system is preferred over the baseline in the majority of the cases for joy, anger and sadness, while the baseline is preferred in the majority of cases for neutral and surprise. This is likely caused by the baseline always speaking in the neutral emotion, since it lacks a conditioning signal. For the surprise emotion, the reason of the preference for the baseline is less clear. It is likely due to the surprised samples of our proposed system having extreme prosodic variation, which could be seen as generally unnatural and therefore not fitting for any presented utterance.  

%\begin{figure}
%	\centering
%	\includegraphics[width=0.9\columnwidth, clip, trim=0.4cm 5.8cm 0.8cm 5.7cm]{figures/pref_total_pie.pdf}
%	\caption{Results of preference test. Prompt Conditioned TTS is preferred in 54.3\% of test cases.}
%	\label{fig:pref_total}
%\end{figure}

%\begin{figure}
%	\centering
%	\includegraphics[width=0.9\columnwidth, clip, trim=5cm 3.2cm 2.4cm 2.2cm]{figures/pref_pie.pdf}
%	\caption{Results of preference test. Prompt Conditioned TTS is preferred in 54.3\% of all test cases. The baseline was preferred in 34.4\% of all cases.}
%	\label{fig:pref_total}
%\end{figure}

\subsubsection{Emotional Style Transfer}
Finally, the participants are presented synthesized speech from the prompt conditioned system where the same prompt is used for multiple utterances with mismatching emotional content and are asked to rate the similarity of the speech samples to the prompt with respect to their prosodic realization on a 5-point scale. We received 320 prosody similarity ratings. The results are displayed in Table~\ref{tab:similarity}. The overall strong ratings for both speakers demonstrate that the model accurately follows the prompt for realizing speech emotion and that this emotion can effectively be transferred to arbitrary utterances, even those with different emotional content, by using the same prompt.

\begin{table}
    \begin{center}
        \begin{tabular}{l|cc}
            \toprule
            Emotion & Female (speaker 15) & Male (speaker 14) \\
            \midrule
            anger & $4.61 \pm 1.63$ & $4.12 \pm 1.73$ \\
            joy & $3.30 \pm 1.60$ & $3.63 \pm 1.66$ \\
            neutral & $4.52 \pm 1.67$ & $4.41 \pm 1.65$ \\
            sadness & $4.35 \pm 1.68$ & $4.66 \pm 1.60$ \\
            surprise & $4.22 \pm 1.42$ & $4.55 \pm 1.66$ \\
            \midrule
            Overall & $4.19 \pm 0.94$ & $4.27 \pm 0.89$ \\
            \bottomrule
        \end{tabular}
    \end{center}
    \caption{Mean similarity ratings (5-point scale) across all emotions and two speakers. Utterances were produced with the same prompt but different input text. The speaker IDs refer to the ones in ESD.}
    \label{tab:similarity}
\end{table}

\section{Conclusion}
In this work, we propose a text-to-speech system that is conditioned on embeddings extracted from natural language prompts which makes the prosodic parameters of generated speech controllable in an intuitive and effective way. Prompt embeddings are concatenated with speaker embeddings and provided as input to the model's encoder, decoder and prosody predictors. Furthermore the proposed training strategy merges emotional speech and text datasets to obtain relevant prompts which are varied in each iteration, increasing the generalization capability and reducing the risk of overfitting. Evaluation results confirm the prosodic controllability through prompting while maintaining high speech quality and multi-speaker capability.

%\section{Limitations \& Future Work}
%Due to the use of a speaker embedding matrix, the model is not capable of synthesizing utterances in the voice of speakers unseen during training. %The main challenge hereby one would have to use pre-trained speaker embeddings instead which contain speaking style related information in addition to speaker timbre. This has an impact on the prosody prediction of the model and dampens the effect of the prompt embedding. In future work it could be explored if techniques for disentangling speaker timbre from prosodic information in speaker embeddings such as in \cite{An2022} provide applicable solutions.
%Another current limitation is that the emotional intensity cannot be controlled. The emotional speech datasets however provide annotation for utterances with different intensities which could be leveraged. The problem hereby would be to obtain corresponding prompts with different peculiarities.

\bibliographystyle{IEEEtran}
\bibliography{mybib}

\end{document}